\documentclass[runningheads]{llncs}
\usepackage{graphicx}
\usepackage{hyperref}
\usepackage{url}
\usepackage{subfigure}
\usepackage{multirow}
\usepackage{algorithm}
\usepackage{algorithmic}
\usepackage{color}
\usepackage{amsmath}
\usepackage{booktabs}
\begin{document}
\title{An Empirical Study on Feature Discretization}
\author{Qiang Liu\inst{1,2} \and
Zhaocheng Liu\inst{1} \and
Haoli Zhang\inst{1}}
\authorrunning{Liu et al.}
\institute{RealAI, Beijing, China \and
Tsinghua University, Beijing, China
\email{\{qiang.liu,zhaocheng.liu,haoli.zhang\}@realai.ai}}
\maketitle              % typeset the header of the contribution
\begin{abstract}
When dealing with continuous numeric features, we usually adopt feature discretization.
In this work, to find the best way to conduct feature discretization, we present some theoretical analysis, in which we focus on analyzing correctness and robustness of feature discretization.
Then, we propose a novel discretization method called Local Linear Encoding (LLE).
Experiments on two numeric datasets show that, LLE can outperform conventional discretization method with much fewer model parameters.
\end{abstract}
\section{Introduction}

In various machine learning tasks, feature discretization has long been a commonly-used trick for dealing with numeric features.
Meanwhile, feature discretization has been proven useful to improve the capability of numerical features \cite{liu2002discretization,kotsiantis2006discretization,chapelle2015simple,yuanfei2019autocross}.
We can conduct feature discretization on numerical feature fields to generate corresponding categorical feature fields, and then perform variety of machine learning models, e.g., Deep Neural Networks (DNN) or Logistic Regression (LR).
Usually, we conduct equal-frequency discretization or equal-value discretization.
Previous research works \cite{fayyad1993multi,clarke2000entropy,liu2002discretization,kotsiantis2006discretization,franc2018learning} mainly focus on determining the optimal boundaries of discrete bins.
In \cite{yuanfei2019autocross}, an approach named Multi-Granularity Discretization (MGD) is proposed, where granularity means the number of discrete bins.
Instead of using a fine-tuned granularity, MGD discretizes each numeric feature field into several, rather than only one, categorical feature fields, each with a different granularity.
Then, MGD evaluates the performances of features with different granularities on the validation set, and keeps only the best half.

To find the best way to conduct feature discretization, we present some theoretical analysis on the correctness and robustness of feature discretization.
Then, we propose a novel discretization method called Local Linear Encoding (LLE).
In conventional feature discretization, we assign one embedding for each discrete bin when performing sparse DNN or sparse LR.
In contrast, with LLE, we conduct linear interpolation in each discrete bin.
LLE can improve the correctness of feature discretization, while preserving the robustness.
Experiments on two numeric datasets show that, LLE can outperform conventional discretization method with much fewer model parameters.

\section{Theoretical Analysis on Feature Discretization}

In this subsection, we are going to conduct some theoretical analysis on feature discretization, which investigates how good a discrete bin is for representing original continuous numeric features.
Suppose we have continuous numeric features ${\bf{v}} = \{ {\mathop v\nolimits_1 ,\mathop v\nolimits_2 ,...,\mathop v\nolimits_{\left| {\bf{v}} \right|} } \}$ in the corresponding field,
the ground-truth projection from the features to the labels is $\lambda\left( {\bf{v}} \right)$,
we assume the observations are under normal distribution ${\bf{O}} \sim N\left( {\lambda \left( {\bf{v}} \right),\mathop \sigma \nolimits^2 } \right)$,
and the corresponding predictions after discretization are denoted as ${\bf{P}}$.
Moreover, we have a discrete bin $B$ with features $\mathop {\bf{v}}\nolimits_B  = \left\{ {\mathop v\nolimits_a ,...,\mathop v\nolimits_b } \right\}$, whose lower boundary is $\mathop v\nolimits_a$ and upper boundary is $\mathop v\nolimits_b$.
In the common feature discretization, for any feature ${\mathop v\nolimits_i  \in \mathop {\bf{v}}\nolimits_B }$, we have
\begin{equation} \label{equation:common_bin}
\mathop p\nolimits_i  = \frac{1}{{\left| B \right|}}\sum\limits_{\mathop v\nolimits_j  \in \mathop {\bf{v}}\nolimits_B } {\mathop o\nolimits_j }.
\end{equation}
We analyze the bin $B$ from two perspectives: correctness and robustness.
Correctness means whether the discrete bin can correctly represent original continuous numeric features.
Robustness verifies whether the discrete bin is robust to noise.
We define the evaluation of correctness and robustness as follows.

\begin{definition}(Correctness of A Discrete Bin) \label{def:correctness}
For a discrete bin $B$, the correctness is the expectation of the error between the ground-truth labels and the predictions
\begin{equation} \label{equation:correctness}
{\rm{Correctness}}\left( B \right) = \mathop E\nolimits_{\mathop {\bf{O}}\nolimits_B  \sim N\left( {\lambda \left( {\mathop {\bf{v}}\nolimits_B } \right),\mathop \sigma \nolimits^2 } \right)} \left[ {\frac{1}{{\left| B \right|}}\sum\limits_{\mathop v\nolimits_i  \in \mathop {\bf{v}}\nolimits_B } {\mathop {\left( {\lambda \left( {\mathop v\nolimits_i } \right) - \mathop p\nolimits_i } \right)}\nolimits^2 } } \right],
\end{equation}
which is the smaller, the better.
\end{definition}

\begin{definition}(Robustness of A Discrete Bin) \label{def:robustness}
For a discrete bin $B$, due to the noise in the observations, the robustness is the variance of the predictions
\begin{equation} \label{equation:robustness}
{\rm{Robustness}}\left( B \right) = \mathop V\nolimits_{\mathop {\bf{O}}\nolimits_B  \sim N\left( {\lambda \left( {\mathop {\bf{v}}\nolimits_B } \right),\mathop \sigma \nolimits^2 } \right)} \left[ {\frac{1}{{\left| B \right|}}\sum\limits_{\mathop v\nolimits_i  \in \mathop {\bf{v}}\nolimits_B } {\mathop p\nolimits_i } } \right],
\end{equation}
which is the smaller, the better.
\end{definition}

In our common sense, when we have more samples in a bin, i.e., we have less bins in total, the correctness will be worse, and the robustness will be better.

\begin{lemma} \label{lemma:Correctness}
When we have less samples in a discrete bin, i.e., ${\left| B \right|}$ is smaller, the corresponding correctness of the bin will be better, i.e., ${\rm{Correctness}}\left( B \right)$ will be smaller.
\end{lemma}

\emph{Proof}. Combing Eq. (\ref{equation:common_bin}) and Eq. (\ref{equation:correctness}), we have
\begin{equation}
\begin{array}{l}
{\rm{Correctness}}\left( B \right) = \mathop E\nolimits_{\mathop {\bf{O}}\nolimits_B  \sim N\left( {\lambda \left( {\mathop {\bf{v}}\nolimits_B } \right),\mathop \sigma \nolimits^2 } \right)} \left[ {\frac{1}{{\left| B \right|}}\sum\limits_{\mathop v\nolimits_i  \in \mathop {\bf{v}}\nolimits_B } {\mathop {\left( {\lambda \left( {\mathop v\nolimits_i } \right) - \frac{1}{{\left| B \right|}}\sum\limits_{\mathop v\nolimits_j  \in \mathop {\bf{v}}\nolimits_B } {\mathop o\nolimits_j } } \right)}\nolimits^2 } } \right]\\
\quad\quad\quad\quad\quad\quad\quad = \frac{1}{{\left| B \right|}}\sum\limits_{\mathop v\nolimits_i  \in \mathop {\bf{v}}\nolimits_B } {\mathop {\left( {\lambda \left( {\mathop v\nolimits_i } \right) - \frac{1}{{\left| B \right|}}\sum\limits_{\mathop v\nolimits_j  \in \mathop {\bf{v}}\nolimits_B } {\lambda \left( {\mathop v\nolimits_j } \right)} } \right)}\nolimits^2 }  = \mathop V\nolimits_{\mathop v\nolimits_i  \in \mathop {\bf{v}}\nolimits_B } \left[ {\lambda \left( {\mathop v\nolimits_i } \right)} \right]
\end{array}.
\end{equation}
Then, we equally divide the bin $B$ into two bins $B_1$ and $B_2$, i.e., $\left| {\mathop B\nolimits_1 } \right| = \left| {\mathop B\nolimits_2 } \right| < \left| B \right|$.
The overall correctness of $B_1$ and $B_2$ becomes
\begin{equation}
\left( {\mathop V\nolimits_{\mathop v\nolimits_i  \in \mathop {\bf{v}}\nolimits_{\mathop B\nolimits_1 } } \left[ {\lambda \left( {\mathop v\nolimits_i } \right)} \right] + \mathop V\nolimits_{\mathop v\nolimits_i  \in \mathop {\bf{v}}\nolimits_{\mathop B\nolimits_2 } } \left[ {\lambda \left( {\mathop v\nolimits_i } \right)} \right]} \right)/2.
\end{equation}
According to the characteristics of Between Group Variance (BGV), we have
\begin{equation}
\mathop  V\nolimits_{\mathop v\nolimits_i  \in \mathop {\bf{v}}\nolimits_B } \left[ {\lambda \left( {\mathop v\nolimits_i } \right)} \right] = \frac{{\mathop  V\nolimits_{\mathop v\nolimits_i  \in \mathop {\bf{v}}\nolimits_{\mathop B\nolimits_1 } } \left[ {\lambda \left( {\mathop v\nolimits_i } \right)} \right] + \mathop  V\nolimits_{\mathop v\nolimits_i  \in \mathop {\bf{v}}\nolimits_{\mathop B\nolimits_2 } } \left[ {\lambda \left( {\mathop v\nolimits_i } \right)} \right]}}{2} + \mathop {\rm{BGV}}\left[ {\lambda \left( {\mathop {\bf{v}}\nolimits_{\mathop B\nolimits_1 } } \right),\lambda \left( {\mathop {\bf{v}}\nolimits_{\mathop B\nolimits_2 } } \right)} \right].
\end{equation}
When ${\rm{Mean}}\left[ {\lambda \left( {\mathop {\bf{v}}\nolimits_{\mathop B\nolimits_1 } } \right)} \right] = {\rm{Mean}}\left[ {\lambda \left( {\mathop {\bf{v}}\nolimits_{\mathop B\nolimits_2 } } \right)} \right]$, we have $\mathop {\rm{BGV}}\left[ {\lambda \left( {\mathop {\bf{v}}\nolimits_{\mathop B\nolimits_1 } } \right),\lambda \left( {\mathop {\bf{v}}\nolimits_{\mathop B\nolimits_2 } } \right)} \right] = 0$.
Meanwhile, in most cases, we have $\mathop {{\rm{Mean}}}\nolimits_{\mathop v\nolimits_i  \in \mathop {\bf{v}}\nolimits_{\mathop B\nolimits_1 } } \left[ {\lambda \left( {\mathop v\nolimits_i } \right)} \right] \ne {\rm{Mean}}\left[ {\lambda \left( {\mathop {\bf{v}}\nolimits_{\mathop B\nolimits_1 } } \right)} \right]$, and $\mathop {\rm{BGV}}\left[ {\lambda \left( {\mathop {\bf{v}}\nolimits_{\mathop B\nolimits_1 } } \right),\lambda \left( {\mathop {\bf{v}}\nolimits_{\mathop B\nolimits_2 } } \right)} \right] > 0$.
Finally, we can obtain
\begin{equation}
{\rm{Correctness}}\left( B \right) \ge \frac{{{\rm{Correctness}}\left( {\mathop B\nolimits_1 } \right) + {\rm{Correctness}}\left( {\mathop B\nolimits_1 } \right)}}{2},
\end{equation}
which proves Lemma (\ref{lemma:Correctness}).

\begin{lemma} \label{lemma:robustness}
When we have more samples in a discrete bin, i.e., ${\left| B \right|}$ is larger, the corresponding robustness of the bin will be better, i.e., ${\rm{Robustness}}\left( B \right)$ will be smaller.
\end{lemma}

\emph{Proof}. First, we need to calculate $\mathop  V\nolimits_{\mathop {\bf{O}}\nolimits_B  \sim N\left( {\lambda \left( {\mathop {\bf{v}}\nolimits_B } \right),\mathop \sigma \nolimits^2 } \right)} \left[ {\mathop p\nolimits_i } \right]$ as
\begin{equation}
\small
\mathop  V\nolimits_{\mathop {\bf{O}}\nolimits_B  \sim N\left( {\lambda \left( {\mathop {\bf{v}}\nolimits_B } \right),\mathop \sigma \nolimits^2 } \right)} \left[ {\mathop p\nolimits_i } \right] = \mathop  V\nolimits_{\mathop {\bf{O}}\nolimits_B  \sim N\left( {\lambda \left( {\mathop {\bf{v}}\nolimits_B } \right),\mathop \sigma \nolimits^2 } \right)} \left[ {\frac{1}{{\left| B \right|}}\sum\limits_{\mathop v\nolimits_j  \in \mathop {\bf{v}}\nolimits_B } {\mathop o\nolimits_j } } \right] = \sum\limits_{\mathop v\nolimits_j  \in \mathop {\bf{v}}\nolimits_B } {\mathop {\left( {\frac{1}{{\left| B \right|}}} \right)}\nolimits^2 \mathop \sigma \nolimits^2 }  = \frac{{\mathop \sigma \nolimits^2 }}{{\left| B \right|}}.
\end{equation}
Then, Eq. (\ref{equation:robustness}) can be rewritten as 
\begin{equation} \label{equation:robustness_LD}
{\rm{Robustness}}\left( B \right) = \sum\limits_{\mathop v\nolimits_j  \in \mathop {\bf{v}}\nolimits_B } {\mathop {\left( {\frac{1}{{\left| B \right|}}} \right)}\nolimits^2 \frac{{\mathop \sigma \nolimits^2 }}{{\left| B \right|}}}  = \frac{{\mathop \sigma \nolimits^2 }}{{\mathop {\left| B \right|}\nolimits^2 }},
\end{equation}
which proves Lemma (\ref{lemma:robustness}).

\section{Local Linear Encoding}

In this subsection, we are going to find a suitable discretization approach with better correctness.
Inspired by some previous works on dealing with continuous time values \cite{liu2016predicting,liu2016context,liu2017multi,liu2018mining,yu2019attention}, we propose a novel feature discretization method called Local Linear Encoding (LLE).
In conventional feature discretization, we assign one embedding for each discrete bin when performing sparse DNN or sparse LR.
In contrast, with LLE, we conduct linear interpolation in each discrete bin.
Specifically, in the bin $B$, for a numeric feature $v_c$, we can lookup its embedding in sparse LR or sparse DNN as
\begin{equation}
\mathop e\nolimits_c  = \alpha \mathop e\nolimits_a  + \beta \mathop e\nolimits_b,
\end{equation}
where $\mathop e\nolimits_a$ and $\mathop e\nolimits_b$ are the embeddings for the lower boundary and the upper boundary of the bin $B$, and the two weights can be calculated as
\begin{equation}
\alpha  = \frac{{\mathop v\nolimits_b  - \mathop v\nolimits_c }}{{\mathop v\nolimits_b  - \mathop v\nolimits_a }},
\end{equation}
\begin{equation}
\beta  = \frac{{\mathop v\nolimits_c  - \mathop v\nolimits_a }}{{\mathop v\nolimits_b  - \mathop v\nolimits_a }}.
\end{equation}

Then, we need to analyze whether LLE is better than common feature discretization.
Regarding the same discrete bin $B$, compared with common feature discretization, LLE should have better correctness and similar robustness.

\begin{lemma} \label{lemma:LLE_Correctness}
Regarding the same bin $B$, LLE has better correctness than common feature discretization.
\end{lemma}

\emph{Proof}. Via linear interpolation in the bin, we can obtain
\begin{equation}
\mathop {\bf{P}}\nolimits_B  = \mathop {\bf{v}}\nolimits_B {\bf{\hat W}}  + \hat b.
\end{equation}
According to the absolute solution of linear regression, we have
\begin{equation} \label{equation:LLE_W}
{\bf{\hat W}} = \frac{{\sum\limits_{\mathop v\nolimits_j  \in \mathop {\bf{v}}\nolimits_B } {\mathop v\nolimits_j \mathop o\nolimits_j } }}{{\sum\limits_{\mathop v\nolimits_j  \in \mathop {\bf{v}}\nolimits_B } {\mathop v\nolimits_j^2 } }},
\end{equation}
\begin{equation}
\hat b = \frac{1}{{\left| B \right|}}\sum\limits_{\mathop v\nolimits_j  \in \mathop {\bf{v}}\nolimits_B } {\left( {\mathop o\nolimits_j  - \mathop v\nolimits_j {\bf{\hat W}}} \right)},
\end{equation}
from which we have
\begin{equation}
\mathop  E\nolimits_{\mathop {\bf{O}}\nolimits_B  \sim N\left( {\lambda \left( {\mathop {\bf{v}}\nolimits_B } \right),\mathop \sigma \nolimits^2 } \right)} \left[ {{\bf{\hat W}}} \right] = \frac{{\sum\limits_{\mathop v\nolimits_j  \in \mathop {\bf{v}}\nolimits_B } {\mathop v\nolimits_j \lambda\left( {\mathop v\nolimits_j } \right)} }}{{\sum\limits_{\mathop v\nolimits_j  \in \mathop {\bf{v}}\nolimits_B } {\mathop v\nolimits_j^2 } }},
\end{equation}
\begin{equation}
\mathop  E\nolimits_{\mathop {\bf{O}}\nolimits_B  \sim N\left( {\lambda \left( {\mathop {\bf{v}}\nolimits_B } \right),\mathop \sigma \nolimits^2 } \right)} \left[ {\hat b} \right] = \frac{1}{{\left| B \right|}}\sum\limits_{\mathop v\nolimits_j  \in \mathop {\bf{v}}\nolimits_B } {\left( {\lambda\left( {\mathop v\nolimits_j } \right) - \mathop v\nolimits_j \mathop E\nolimits_{\mathop {\bf{O}}\nolimits_B  \sim N\left( {\lambda \left( {\mathop {\bf{v}}\nolimits_B } \right),\mathop \sigma \nolimits^2 } \right)} \left[ {{\bf{\hat W}}} \right]} \right)}.
\end{equation}
Then, Eq. (\ref{equation:correctness}) can be rewritten as
\begin{equation}
\scriptsize
\begin{array}{l}
{\rm{Correctness}}\left( B \right) = \mathop  E\nolimits_{\mathop {\bf{O}}\nolimits_B  \sim N\left( {\lambda \left( {\mathop {\bf{v}}\nolimits_B } \right),\mathop \sigma \nolimits^2 } \right)} \left[ {\frac{1}{{\left| B \right|}}\sum\limits_{\mathop v\nolimits_i  \in \mathop {\bf{v}}\nolimits_B } {\mathop {\left( {\lambda \left( {\mathop v\nolimits_i } \right) - \mathop {\bf{v}}\nolimits_B {\bf{\hat W}} - \hat b} \right)}\nolimits^2 } } \right]\\
\quad\quad\quad\quad\quad\quad\quad = \frac{1}{{\left| B \right|}}\sum\limits_{\mathop v\nolimits_i  \in \mathop {\bf{v}}\nolimits_B } {\mathop {\left( {\left( {\lambda \left( {\mathop v\nolimits_i } \right) - \frac{1}{{\left| B \right|}}\sum\limits_{\mathop v\nolimits_j  \in \mathop {\bf{v}}\nolimits_B } {\lambda \left( {\mathop v\nolimits_j } \right)} } \right) - \frac{{\sum\limits_{\mathop v\nolimits_j  \in \mathop {\bf{v}}\nolimits_B } {\mathop v\nolimits_j \lambda \left( {\mathop v\nolimits_j } \right)} }}{{\sum\limits_{\mathop v\nolimits_j  \in \mathop {\bf{v}}\nolimits_B } {\mathop v\nolimits_j^2 } }}\left( {\mathop v\nolimits_j  - \frac{1}{{\left| B \right|}}\sum\limits_{\mathop v\nolimits_j  \in \mathop {\bf{v}}\nolimits_B } {\mathop v\nolimits_j } } \right)} \right)}\nolimits^2 } \\
\quad\quad\quad\quad\quad\quad\quad \le \frac{1}{{\left| B \right|}}\sum\limits_{\mathop v\nolimits_i  \in \mathop {\bf{v}}\nolimits_B } {\mathop {\left( {\lambda \left( {\mathop v\nolimits_i } \right) - \frac{1}{{\left| B \right|}}\sum\limits_{\mathop v\nolimits_j  \in \mathop {\bf{v}}\nolimits_B } {\lambda \left( {\mathop v\nolimits_j } \right)} } \right)}\nolimits^2 } 
\end{array}.
\end{equation}

\begin{table}[htbp]
  \centering
  \label{tab:performance}
  \caption{Performance comparison of different feature discretization approaches}
    \begin{tabular}{c|c|ccc|ccc}
    \toprule
    \multirow{2}[2]{*}{model} & \multirow{2}[2]{*}{feature discretization approach} & \multicolumn{3}{c|}{HIGGS} & \multicolumn{3}{c}{SUSY} \\
          &       & 100\% & 10\%  & 1\%   & 100\% & 10\%  & 1\% \\
    \midrule
    \multirow{7}[2]{*}{LR} & CD (5) & 74.99 & 75    & 74.93 & 85.85 & 85.84 & 85.62 \\
          & CD (10) & 76.42 & 76.18 & 76.26 & 86.69 & 86.57 & 85.96 \\
          & CD (100) & 77.29 & 77.26 & 75.84 & 87.06 & 86.17 & 83.72 \\
          & MGD   & 77.32 & 77.29 & 76.48 & 87.08 & 86.71 & 85.78 \\
          & LLE (5) & 77.07 & 77.05 & 76.95 & 86.77 & 86.47 & 85.98 \\
          & LLE (10) & 77.42 & \textbf{77.38} & \textbf{77.21} & \textbf{87.13} & \textbf{86.83} & \textbf{86.18} \\
          & LLE (100) & \textbf{77.46} & 77.31 & 75.89 & 87.08 & 86.23 & 83.96 \\
    \midrule
    \multirow{7}[2]{*}{DNN} & CD (5) & 79.42 & 78.9  & 77.92 & 86.68 & 86.08 & 85.75 \\
          & CD (10) & 81.67 & 81.07 & 79.49 & 87.28 & 86.77 & 82.49 \\
          & CD (100) & 82.59 & 81.09 & 76.93 & 87.41 & 77.25 & 82.47 \\
          & MGD   & 82.76 & 81.31 & 79.68 & 87.43 & 86.66 & 84.95 \\
          & LLE (5) & 82.68 & 82.16 & \textbf{80.43} & 87.53 & 86.83 & 86.23 \\
          & LLE (10) & \textbf{82.87} & \textbf{82.24} & 80.35 & \textbf{87.58} & \textbf{87.08} & \textbf{86.62} \\
          & LLE (100) & 82.82 & 81.39 & 77.21 & 87.46 & 78.62 & 82.61 \\
    \bottomrule
    \end{tabular}%
\end{table}%

\begin{lemma} \label{lemma:LLE_Robustness}
Regarding the same bin $B$, LLE has the same robustness as common feature discretization.
\end{lemma}

\emph{Proof}. According to Eq. (\ref{equation:LLE_W}), we have
\begin{equation}
\mathop  V\nolimits_{\mathop {\bf{O}}\nolimits_B  \sim N\left( {\lambda \left( {\mathop {\bf{v}}\nolimits_B } \right),\mathop \sigma \nolimits^2 } \right)} \left[ {{\bf{\hat W}}} \right] = \sum\limits_{\mathop v\nolimits_i  \in \mathop {\bf{v}}\nolimits_B } {\mathop {\left( {\frac{{\mathop v\nolimits_i }}{{\sum\limits_{\mathop v\nolimits_j  \in \mathop {\bf{v}}\nolimits_B } {\mathop v\nolimits_j^2 } }}} \right)}\nolimits^2 \mathop \sigma \nolimits^2  = \frac{{\sum\limits_{\mathop v\nolimits_j  \in \mathop {\bf{v}}\nolimits_B } {\mathop v\nolimits_j^2 } }}{{\mathop {\left( {\sum\limits_{\mathop v\nolimits_j  \in \mathop {\bf{v}}\nolimits_B } {\mathop v\nolimits_j^2 } } \right)}\nolimits^2 }}\mathop \sigma \nolimits^2  = } \frac{{\mathop \sigma \nolimits^2 }}{{\sum\limits_{\mathop v\nolimits_j  \in \mathop {\bf{v}}\nolimits_B } {\mathop v\nolimits_j^2 } }}.
\end{equation}
Then, Eq. (\ref{equation:robustness}) can be rewritten as
\begin{equation}
\begin{array}{l}
{\rm{Robustness}}\left( B \right) = \mathop  V\nolimits_{\mathop {\bf{O}}\nolimits_B  \sim N\left( {\lambda \left( {\mathop {\bf{v}}\nolimits_B } \right),\mathop \sigma \nolimits^2 } \right)} \left[ {\frac{1}{{\left| B \right|}}\sum\limits_{\mathop v\nolimits_i  \in \mathop {\bf{v}}\nolimits_B } {\left( {\mathop v\nolimits_i {\bf{\hat W}} + \hat b} \right)} } \right]\\
\quad\quad\quad\quad\quad\quad\ = \sum\limits_{\mathop v\nolimits_i  \in \mathop {\bf{v}}\nolimits_B } {\mathop {\left( {\frac{{\mathop v\nolimits_i }}{{\left| B \right|}}} \right)}\nolimits^2 \frac{{\mathop \sigma \nolimits^2 }}{{\sum\limits_{\mathop v\nolimits_j  \in \mathop {\bf{v}}\nolimits_B } {\mathop v\nolimits_j^2 } }}}  = \frac{{\sum\limits_{\mathop v\nolimits_j  \in \mathop {\bf{v}}\nolimits_B } {\mathop v\nolimits_j^2 } }}{{\mathop {\left| B \right|}\nolimits^2 }}\frac{{\mathop \sigma \nolimits^2 }}{{\sum\limits_{\mathop v\nolimits_j  \in \mathop {\bf{v}}\nolimits_B } {\mathop v\nolimits_j^2 } }} = \frac{{\mathop \sigma \nolimits^2 }}{{\mathop {\left| B \right|}\nolimits^2 }}
\end{array},
\end{equation}
which is the same as in Eq. (\ref{equation:robustness_LD}).

\section{Experiments}

In this section, we evaluate the performances of our proposed LLE approach on feature discretization. 

\begin{table}[htbp]
  \centering
  \caption{The count of model parameters.}
    \begin{tabular}{c|ccc|ccc}
    \toprule
    \multirow{2}[2]{*}{feature discretization approach} & \multicolumn{3}{c|}{HIGGS} & \multicolumn{3}{c}{SUSY} \\
          & 100\% & 10\%  & 1\%   & 100\% & 10\%  & 1\% \\
    \midrule
    MGD   & 162560 & 148140 & 141680 & 116820 & 41580 & 21690 \\
    LLE   & 308   & 308   & 168   & 198   & 198   & 198 \\
    \bottomrule
    \end{tabular}%
  \label{tab:count}%
\end{table}%

\subsection{Experimental Settings}

We compare LLE with two baselines: Common Discretization (CD) and Multi-Granularity Discretization (MGD).
The granularities in MGD are searched in $\{ 10,100,1000,10000 \}$.
We run CD and LLE with three different numbers of discrete bins: $5$, $10$ and $100$.
Meanwhile, all the discretization approaches are performed with both LR and DNN.
The datasets we adopt are two numeric datasets: HIGGS\footnote{\url{https://archive.ics.uci.edu/ml/datasets/HIGGS}} and SUSY\footnote{\url{http://archive.ics.uci.edu/ml/datasets/SUSY}}.

\subsection{Experimental Analysis}

In Tab. (\ref{tab:performance}), we illustrate the performances of different feature discretization approaches with different ratios of training samples, in which we perform both LR and DNN.
It is clear that, with different ratios of training samples, the best number in common discretization varies a lot.
Meanwhile, MGD can search for a relative suitable granularity, and achieve better performances compared with CD.
Moreover, LLE performs the best on all datasets with all ratios of training samples.
And with different ratios of training samples, the performances of LLE are similar.

Moreover, Tab. (\ref{tab:count}) shows the count of mode parameters when performing MGD and LLE.
Obviously, compared with MGD, LLE can achieve better performances with much less model parameters.

\section{Conclusion}

When performing variety of machine learning models, we usually need feature discretization for preprocessing.
To find the best way to conduct feature discretization, we first present some theoretical analysis on feature discretization, and then propose a novel discretization method called Local Linear Encoding (LLE).
Experiments on two numeric datasets show that, LLE can outperform conventional discretization method with much fewer model parameters.

\bibliographystyle{splncs}
\bibliography{LLE}

\end{document}